\documentclass[sigconf,balance=false]{acmart}
\usepackage{popets}
\usepackage{colortbl}
\usepackage[justification=centering]{caption}

\setcopyright{popets}
\copyrightyear{YYYY}

\acmYear{2024}
\acmVolume{YYYY}
\acmNumber{X}
\acmDOI{XXXXXXX.XXXXXXX}
\acmISBN{}
\acmConference{Proceedings on Privacy Enhancing Technologies}
\begin{document}

\title{SoK: Reducing the Vulnerability of Fine-tuned Language Models to Membership Inference Attacks}


\author{Guy Amit}
\orcid{1234-5678-9012}
\affiliation{%
  \institution{IBM Research}
  \city{Haifa}
  \country{Israel}}
\email{guy.amit@ibm.com}

\author{Abigail Goldsteen}
\orcid{1234-5678-9012}
\affiliation{%
  \institution{IBM Research}
  \city{Haifa}
  \country{Israel}}
\email{abigailt@il.ibm.com}

\author{Ariel Farkash}
\orcid{1234-5678-9012}
\affiliation{%
  \institution{IBM Research}
  \city{Haifa}
  \country{Israel}}
\email{arielf@il.ibm.com}


\renewcommand{\shortauthors}{Amit et al.}

\begin{abstract}
Natural language processing models have experienced a significant upsurge in recent years, with numerous applications being built upon them. Many of these applications require fine-tuning generic base models on customized, proprietary datasets. This fine-tuning data is especially likely to contain personal or sensitive information about individuals, resulting in increased privacy risk. Membership inference attacks are the most commonly employed attack to assess the privacy leakage of a machine learning model. However, limited research is available on the factors that affect the vulnerability of language models to this kind of attack, or on the applicability of different defense strategies in the language domain. We provide the first systematic review of the vulnerability of fine-tuned large language models to membership inference attacks, the various factors that come into play, and the effectiveness of different defense strategies. We find that some training methods provide significantly reduced privacy risk, with the combination of differential privacy and low-rank adaptors achieving the best privacy protection against these attacks.
\end{abstract}


\maketitle

\section{Introduction}
The utilization of Natural Language Processing (NLP) models has experienced a significant upsurge in recent years, with numerous applications being built upon these models~\cite{min2023recent}, including chatbots, sentiment analysis, and automatic code generation tools. Many of these applications require fine-tuning generic base models on customized and proprietary datasets. 
In some cases, this data may contain personal or sensitive information about individuals, which could result in increased privacy risk. Regulations such as the General Data Protection Regulation (GDPR)\footnote{\url{https://gdpr.eu/}} and the California Consumer Privacy Act (CCPA)\footnote{\url{https://oag.ca.gov/privacy/ccpa}} have been implemented in Europe and the Unites States respectively in an attempt to govern and safeguard the use of personal information. 

When a Machine Learning (ML) model is trained, knowledge from the training dataset is infused into it. In fact, part of the dataset may be memorized by the model. In other words, instead of learning general trends and patterns that characterize any data drawn from the same distribution, the model inadvertently learns detailed characteristics that are unique to the specific training samples to which it was exposed, sometimes even memorizing complete training samples. This gives way to a variety of attacks against ML models, and Large Language Models (LLMs) in particular, that try to take advantage of this phenomenon to infer or extract private information from deployed models. This can include both personal information about individuals, as well as confidential organizational information.

A \textit{Membership inference attack} (MIA) is one such attack that attempts to extract information from an ML model about its training or fine-tuning data. Using simple queries, this attack allows an adversary to infer if a specific data sample was part of a model's training set. If such an attack succeeds, it is typically considered a privacy violation, since private information about individuals who contributed data to the training set can be derived. 
For example, an adversary may use a MIA to deduce that a public figure suffers from a certain disease, and use this information to interfere with their career or even blackmail them~\cite{hu2022membership, shokri2017membership}. MIAs can be used in many domains, such as user behavior analysis in social media~\cite{liu2019socinf}, and detecting credit scores or mortgage returns in banking~\cite{truex2018towards}. The assumption behind most existing MIAs is that models output abnormally high confidence when queried on their training data.

There are a few factors that have been shown to affect the vulnerability of a model to membership inference attacks, most of which can be traced back to the training process. 
These typically include model overfitting, model size and capacity, repeated exposure of the model to the same data samples, and the presence of outliers.
Previous research was mostly focused on classic ML/Deep Learning (DL) models for vision tasks.
With the proliferation of LLMs, we seek to revisit these (and new) factors to better understand their relevance. 

Training LLMs typically involves two distinct phases: pre-training transformers as general-purpose language models (such as Roberta~\cite{liu2019roberta} or GPT2~\cite{radford2019language}) that learn the basic concepts of the underlying language; and fine-tuning them to solve a specific, focused task such as sentiment analysis in movie reviews.
Pre-training is largely performed on publicly available datasets, and the data used to train a model is often also public knowledge.
Fine-tuning is typically performed on a smaller, proprietary dataset, which makes this phase an appropriate target for MIA. Different adaptations made to the fine-tuning process to support large model architectures, such parameter-efficient fine-tuning methods, may also affect the susceptibility of the fine-tuned model to MIA.

In parallel to the efforts to better understand and improve MIA implementations, there is an increasing motivation to try to prevent them by applying mitigation measures to models.
Some of these defense techniques alter the model's training process~\cite{abadi2016deep, wang2022analyzing} and some suggest modifications to the model outputs~\cite{jia2019memguard}. 
Similarly, mitigation techniques have been researched for a long time in the general context of ML, but there is limited research on their effectiveness in the language domain.
Due to the shear size of these models, some defense strategies become impractical or even completely infeasible. We therefore focus on defense mechanisms that can feasibly be applied to LLMs, and specifically fine-tuned LLMs.

In this paper, we perform a comprehensive comparative analysis of the susceptibility of fine-tuned LLMs to membership inference attacks, systematically analyzing various factors that affect their vulnerability to MIA. 
We examine the applicability of existing mitigation techniques to these LLMs , along with several new approaches from recent literature, as well as combinations of these approaches, to determine their effectiveness in preventing MIA.
Our results show that while the susceptibility of undefended or improperly trained LLMs to these attack is relatively high, some of these mitigation methods are able to significantly reduce the privacy risk.

\textbf{Our main contributions are:}
\begin{enumerate}
    \item A first review of membership inference attack vulnerability focused on fine-tuned LLMs.
    \item A systematic review of the different factors that affect the vulnerability of fine-tuned LLMs to membership inference attacks. 
    \item An evaluation of current mitigation techniques and their effectiveness in preventing membership inference attacks in fine-tuned LLMs.
\end{enumerate}

\textbf{Our main findings from this study:}
\begin{enumerate}
\item Most of the mitigation strategies used in traditional ML/DL are also effective when applied to fine-tuning LLMs.
\item Model pruning was found to be relatively ineffective as a defense against MIA.
\item We studied the effect of batch size (a factor that was not yet adequately addressed in prior literature) and found that it has a significant effect on MIA vulnerability, with larger batch sizes providing good protection against this type of attack.
\item The most effective defense strategies for most evaluated models and datasets are differential-privacy based methods such as DP-SGD \cite{abadi2016deep} and  DP-LoRA~\cite{yu2021differentially} (with $\epsilon$=2).
\item A very good accuracy/privacy trade-off can also be achieved just by using LoRA~\cite{hu2021lora} on its own, or in combination with a smaller model size. The LoRA fine-tuning method barely has any negative impact on model accuracy, while providing a significant reduction in MIA risk.
\end{enumerate}

In Section \ref{sec:background}, we survey relevant research in the area of inference attacks against language models. In Section \ref{sec:robust}, we detail various factors that affect the vulnerability of ML models to MIA and present possible defense strategies. Section \ref{sec:setup} describes the experimental setup of this analysis and Section \ref{sec:evaluation} presents its results. Finally, we conclude in Section \ref{sec:conclusion}.

\section{Background}
\label{sec:background}
\subsection{Membership Inference Attacks against Language Models}
\label{sec:MI_attacks}
Membership inference attacks~\cite{shokri2017membership, hu2022membership} are a class of attacks in which an adversary attempts to determine whether a specific data point was used in the training of a target machine learning model.
Formally, for a model $M$ and a training dataset $\mathcal{D}$, the adversary's goal is to determine whether a data point $x$ is part of the training dataset, i.e., $x\in\mathcal{D}$. Samples that were part of the training data are called members, and other samples are non-members.

This kind of attack typically makes use of the model's outputs (called the black-box setting), and operates under the assumption that member samples will cause the model to output abnormal values, such as higher output probabilities. 
Most MIAs employ a binary classification attack model trained to predict the membership status of a data sample based on hand-crafted features produced from the target model's outputs, such as their standard deviation or entropy. 

MIAs are especially relevant when the training dataset comprises sensitive user data or when the model is used to predict sensitive attributes such as medical conditions~\cite{truex2019demystifying, jagannatha2021membership}.
When such an attack succeeds, the adversary may gain insight into private information related to a person, thereby compromising their privacy.
For example, by exposing the participation of an individual in the training dataset used for a loan approval model, an adversary gains knowledge about the individual applying for a loan.

In the context of LLMs, the goal of such an attack is to infer if a specific text segment or document was part of the training dataset used to train (or fine-tune) the model~\cite{he2022membership, shejwalkar2021membership}.
This vulnerability is a significant concern, both because LLMs are widely used in various applications~\cite{min2023recent}, and because they are trained on vast amounts of data, which can be sensitive. 
Any attack that can reveal the membership status of a particular data point can lead to the disclosure of sensitive information, which in turn can have serious consequences on the individuals and organizations that rely on these models. Moreover, this kind of attack can also aid in detecting copyright infringements, and may provide insights on the degree of memorization of a model.

\subsection{Related Surveys}
\begin{table}[h]
\centering
\caption{Papers that perform reviews of privacy attacks}
\label{tab:surveys}
\resizebox{0.45\textwidth}{!}{
\begin{tabular}{l|cccc}
\hline\hline
Paper      &       Attacks    & Defences          & Evaluation             & LLMs                  \\ \hline
 \cite{rigaki2020survey} & $\checkmark $ &  & & \\
 \cite{niu2023sok}  &    $\checkmark$          &        &                      &                      \\
\cite{kuntla2021security} &  $\checkmark$  &           &                      &                      \\
\cite{papernot2018sok} &  $\checkmark$   &    $\checkmark$     &       &          \\
\cite{10.1145/3620667}  &  $\checkmark$    &     $\checkmark$          &         &                      \\
\cite{dionysiou2023sok}  & $\checkmark$   &   $\checkmark$   &  $\checkmark$        &                      \\
\cite{he2022membership} & $\checkmark$ & $\checkmark$ &  $\checkmark$ & Partial  \\ 
Ours & $\checkmark$ & $\checkmark$ &  $\checkmark$ & \checkmark  \\ 

\hline\hline
\end{tabular}
}
\end{table}
In recent years, multiple works have emerged with the primary objective of conducting systematic reviews of privacy attacks on ML models (see Table~\ref{tab:surveys}). 
These surveys cover the definition of MIA and the setups to which MIAs are applicable. 
While certain surveys discuss the subject of MIA mitigation~\cite{papernot2018sok, dionysiou2023sok, he2022membership}, it is noteworthy that only a select few~\cite{he2022membership, dionysiou2023sok} undertake an in-depth comparative analysis to scrutinize this aspect.

The exploration of MIA, and particularly of MIA mitigations, has predominantly centered around the computer vision domain and classical machine learning models.
However, with the prevalence of LLMs, it becomes imperative to extend these works and investigate these topics further through this new lens.
Notably, while the study by He et al.~\cite{he2022membership} discusses the subject of MIA in the context of language models, it falls short in providing a comprehensive evaluation of MIA risks and mitigation strategies tailored specifically for such models.
One exception is the work by Jagannatha et al \cite{jagannatha2021membership}, which examines the MIA vulnerability of differentially private language models in the clinical domain. However, their work does not cover other possible mitigation techniques.

In light of these existing gaps and limitations, this paper aspires to contribute a more thorough comparative analysis addressing the critical domain of MIA mitigation for LLMs.
To the best of our knowledge, this endeavor is a first-of-a-kind effort to tackle this subject matter comprehensively and systematically.

\section{Developing MIA Robust Models}
\label{sec:robust}
Over the years, different techniques have been devised to reduce the privacy risk of ML models.
These techniques are built upon several assumptions regarding different aspects of the training and deployment of these models.
In this section we review some of these techniques and their underlying assumptions.

We start by reviewing the basic factors assumed to affect the vulnerability of ML models to privacy attacks.

\subsection{Factors Affecting MIA Vulnerability}
\label{factors}
Overfitting is often mentioned in the context of MIA as a leading factor contributing to models' vulnerability to these attacks~\cite{shokri2017membership, dionysiou2023sok, he2022membership}. In machine learning, an overfitted model performs accurately on it's training set but is unable to generalize to new data samples. Typically overfitting is measured as the difference between the model's train and test accuracy.

Overfitting can also contribute to the memorization~\cite{carlini2022quantifying} of training data, since when a data sample is presented more times to the model during training than necessary, the chance of it being memorized increases, which also increases the vulnerability to MIA.

Overfitting can be caused by two main factors, (1) over-parametri-zation - too many model trainable parameters compared to the size of the training dataset, and (2) over-training - training the model over too many iterations. In the following we refer to each of these contributing factors separately.\\

\textbf{(1) Model size:} One common assumption about the vulnerability of ML models to MIA, as well as other classes of attacks such as data extraction attacks~\cite{carlini2021extracting, carlini2022quantifying}, is that larger models, i.e., models that contain more parameters, are more capable of memorizing training samples and hence are more vulnerable to attacks.
In the language domain, especially with LLMs used for classification, such as Roberta and Bert, models are often over-parameterized, resulting in perfect accuracy on the training set after only a few training iterations. Not surprisingly, several works focused on large language models~\cite{carlini2021extracting,yu2023bag} have also found that larger models are more prone to memorize their training data and thus are more susceptible to data extraction and membership inference attacks. In our evaluation, we examine various ways of reducing the size of a model and determine how they affect MIA vulnerability.\\

\textbf{(2) Number of training iterations:} As the second factor that can lead to overfitting, we set out to examine different ways in which a model can be over-trained, looking at both the number of training epochs as well as the batch size employed.
We show that when the same model is trained over more iterations, the success of MIA increases. \\

\textbf{(3) Model exposure to training samples:} Several techniques in the field of privacy-preserving ML rely on the assumption that it is less likely for a model to leak information about a sample that was not accurately presented to the model during the training process.
For example, a model that has only processed a sample after some transformation has been applied to it, e.g., rotation for images, will be more robust against MIA~\cite{kaya2021does}. The same is true if the loss gradient used to update the model undergoes some transformation, such as clipping. \\

Taking these factors into consideration, numerous defense methods were invented. Some provide a theoretical privacy guarantee, such as Differential Privacy (DP) based methods~\cite{abadi2016deep}, and some provide practical or empirical protection, striving to reduce the risk of privacy leakage by adapting the models in some way and testing their robustness to attacks. In this paper, we focus on techniques that are applicable to large language models, and specifically to the fine-tuning phase.

\subsection{Differential Privacy for LLMs:}
\textit{Differential Privacy}~\cite{dwork2014algorithmic} is a formal definition of privacy for algorithms that process private datasets, with a privacy parameter (or privacy budget) denoted as $\epsilon$.
An algorithm is said to be differentially private if, by examining it's outputs, one cannot infer if an individual data point was included in the input dataset or not. In other words, DP guarantees that the participation of any individual data point does not have a significant effect on the algorithm outputs. This can be formalized as follows:
for a data processing algorithm $M$, and for every two datasets $\mathcal{D}$ and $\mathcal{D}$' that differ by one data point, the probability of obtaining the same output $x$ from $M$ when used on $\mathcal{D}$ and $\mathcal{D}$' will be different by at most a factor of the exponent of the privacy parameter $\epsilon$:
\begin{equation}
    P(M(\mathcal{D}) = x) \leq e^{\epsilon}\cdot P(M(\mathcal{D}') = x)
\end{equation}

Following this mathematical definition, various algorithms were developed to ensure DP in ML. 
One notable algorithm is \textit{DP-SGD}~\cite{abadi2016deep}, which offers the first application of DP to deep learning models.
\textit{DP-SGD} is a variation of the Stochastic Gradient Descent algorithm~\cite{ruder2016overview} that ensures DP by modifying the gradient updates and limiting the number of training steps.
These two operations help to reduce the second and third factors mentioned above. Specifically, the modification of the gradient updates ensures that the model is never updated according to the original sample, thus protecting it from privacy attacks.

With the advancement of DL, models have become much larger, and it is significantly harder to train these large models using \textit{DP-SGD}. In particular with LLMs, that rely heavily on large transformer-based architectures, it is problematic to apply \textit{DP-SGD} to training without seriously degrading the model's performance~\cite{li2021large}.
This has lead to the development of several DP algorithms more suitable for LLMs~\cite{li2021large, dupuy2022efficient, du2021dp}.

One suggested approach to applying DP to LLMs is to minimize the number of trainable parameters, thus making the application of \textit{DP-SGD} easier.
One such example utilizes LoRA~\cite{hu2021lora}, a technique for fine-tuning (adapting) LLMs for new tasks by inserting adapter layers to the model. Assuming that the rest of the layers were only trained on public data during the pre-training phase, \textit{DP-LoRA}~\cite{yu2021differentially} suggests applying \textit{DP-SGD} to these much smaller adapter layers during the fine-tuning phase, thus facilitating the application of DP to such models.

DP can also be applied to the model outputs~\cite{majmudar2022differentially, ginart2022submix}. 
This may offer protection from MIA in the black-box setting, but it does not guarantee privacy if an adversary has access to the model weights.
The application of these approaches is also only possible for models that output a sequence of tokens (and not classifiers), since the privacy guarantee is built upon repeated sampling from the token distribution. 

\subsection{Empirical Defenses}
While DP offers a mathematically proven privacy defense, researchers have also developed defenses which are not mathematically proven, rather they are empirically tested against actual attack implementations.

One example is \textit{Mem-Guard}~\cite{jia2019memguard} which employs concepts from the field of adversarial ML to modify the model outputs.
The modification to the output is based on an adversarial perturbation~\cite{goodfellow2014explaining} aimed to mislead an attack model (in the black-box setting).
Following \textit{Mem-Guard}, the idea of using adversarial methods was further extended to create a generative-adversarial-network(GAN)-like training process~\cite{goodfellow2020generative} to protect from MIA.
In~\cite{wang2020against}, the target model is trained using a min-max training objective which balances between the model performance and the success of an attack model.
In addition, during the training process, the model weights are pruned (i.e., unnecessary weights are removed from the model), resulting in a smaller model.

In a more recent work~\cite{wang2022analyzing}, the authors also make use of pruning during the fine-tuning process.
In addition to the pruning operarion, which reduces the overall model size, the fine-tuning training objective also enforces that the output probabilities will conform to a uniform distribution, which, according to the authors, improves the model's robustness to MIA.

As previously noted in Section \ref{factors}, overfitting is one of the main factors claimed to contribute to the success of MIAs. 
Building upon this assumption, the authors of~\cite{chen2022relaxloss} suggest a technique to modify the training procedure such that, in addition to minimizing the training loss function, it also includes gradient updates the are aimed at increasing the loss value when under a given threshold.
This technique has been tested on vision models, which are typically trained from scratch, as opposed to LLMs that usually undergo fine-tuning for a specific task. Moreover, the experiments needed to select the threshold value may prove very expensive when applied to LLMs.

\section{Experimental Setup}
\label{sec:setup}
\subsection{Datasets}
In our evaluation, we used two datasets that represent different NLP classification tasks.\\
\begin{itemize}
\item \textbf{Tweet Eval - Hate subset}~\cite{basile-etal-2019-semeval} - A dataset containing 13K tweets from Twitter (now X), each categorized as hateful or not.
\item \textbf{Rotten Tomatoes}~\cite{pang2005seeing} - A dataset containing ~10K movie reviews from the Rotten Tomatoes website. 
Each review is labeled as positive or negative according to the reviewer's opinion of the movie.
\end{itemize}

\subsection{Membership Inference Risk Assessment}
In our evaluation, we tested the model's vulnerability to MIA in a privacy audit or model risk assessment setup. This means that we assumed that an organization is performing the attacks as part of a privacy risk assessment of their own models, and thus has access to the actual fine-tuning dataset. This setup simulates a worst-case scenario, and differs from a malicious attack setup where the adversary is assumed not to have knowledge of the real training data. This audit setup is shared by a few other recent papers \cite{auditing}, \cite{neurons}.

The attack framework employed \cite{framework} is inspired by recent MIA literature~\cite{shokri2017membership, carlini2022membership} and includes multiple input features based on the target model outputs. It employs an exhaustive grid search over different types of attack model architectures, hyper-parameters, input features to the attack model and feature preprocessing methods. It varies ML model implementations (e.g., neural network, random forest, k-nearest neighbors, etc.), input features to the attack (detailed later in this section), and scaling methods (e.g., robust, min-max) to find the combination that achieves the highest attack performance, according to a selected success metric (such as accuracy or Area Under the Receiver Operating Characteristic Curve (AUC-ROC)). 

The input features used by the attack model can include:
\begin{enumerate}
    \item \textbf{Loss} - the cross-entropy loss of the target model outputs.
    \item \textbf{Modified entropy (mentropy)}~\cite{shejwalkar2021membership} - a variation of the entropy of the target model outputs, taking into account the current class label.
    \item \textbf{True Label} - a one-hot encoded representation of the sample's true label. 
    \item \textbf{Predicted Label} - a one-hot encoded representation of the sample's predicted label. 
    \item \textbf{Scaled Class Logits} - the difference between each logit value and the maximal logit value~\cite{carlini2022membership}.
    \item \textbf{Log Likelihood Ratio} - a variation of the Likelihood-ratio Test presented in~\cite{carlini2022membership}, where the mean and standard deviation of the loss are calculated using the training data for the attack model.
\end{enumerate}

To assess the privacy leakage we utilized a test set consisting of an equal number of member and non-member samples for each dataset.

Since the training of some attack models entails randomness, each experiment is repeated 20 times, each with a different random sample of 2000 samples for training the attack model (1000 members and 1000 non-members) and 2000 samples for evaluating the attack (1000 members and 1000 non-members). In each run, the best combination of attack model and additional parameters is selected, and the average metrics across runs are reported, along with their standard deviation (in parentheses in the tables).

\subsection{Evaluation Metrics} 
Throughout our assessment, we utilize three primary performance indicators to evaluate the models: 
\begin{enumerate} 

\item \textbf{Attack ROC curves} - the Receiver Operating Characteristic (ROC) curve represents the relation between the True Positive Rate (TPR) and the False Positive Rate (FPR) of the attack, where FPR measures the proportion of non-member samples incorrectly classified as members, and TPR represents the proportion of member samples correctly identified as members.
Throughout the paper we present the average AUC-ROC (area under the ROC curve), which we use as an aggregate metric for the success of the attacks. The lower this value, the less vulnerable the model is to MIA. For completeness, we also provide full ROC curves for the final comparison between defense methods in Section~\ref{sec:results_summary}. The full ROC curves are taken from the run that yielded the best attack performance (as they cannot be averaged).

\item \textbf{Attack TPR at low FPR} - this metric was proposed by Carlini et al.~\cite{carlini2022membership} and is considered the established way to capture whether an attack can confidently identify members of the training set. We use 1\% as the target FPR. Denoted as TPR@FPR1\%, this metric is calculated as the proportion of correctly identified member samples when the detection threshold is set to predict only 1\% of non-members as members, and is averaged across all 20 runs. The lower this value, the less vulnerable the model is to MIA.


\item \textbf{Model Accuracy} - In addition to the vulnerability to MIA, we evaluate the model's main task performance using standard accuracy, referred to as Model Acc. This gauges the model's ability to accurately execute its designated classification task, where higher Model Acc signifies superior performance of the model.
\end{enumerate}

\subsection{Technical Details} 
All of our evaluations were performed using Nvidia A100 GPUs and pre-trained models from Huggingface~\cite{wolf-etal-2020-transformers} based on PyTorch~\cite{NEURIPS2019_9015}, which we fine-tuned on one of the given datasets. 
In our evaluations, we used the Roberta-base~\cite{liu2019roberta} and Flan-t5-base~\cite{https://doi.org/10.48550/arxiv.2210.11416} models. We trained them using an AdamW optimizer and a linear learning rate scheduler, whose initial value was determined via a search process utilizing the validation set.
The batch size for training was set to $64$ and $128$ for the Flan-t5-base and Roberta-base models respectively. For models making use of \textit{LoRA}~\cite{hu2021lora}, the $R$ parameter was set to $8$ throughout all of our experiments.

\section{Evaluation}
\label{sec:evaluation}
In this section we list the different defense strategies we evaluated, and how each may have contributed to the factors that affect MIA vulnerability. 

\subsection{Number of Training Iterations}
\label{sec:num_iters}

\begin{figure*}
    \centering

    \begin{tabular}{ccc}
         \includegraphics[width=0.65\columnwidth, scale=0.6]{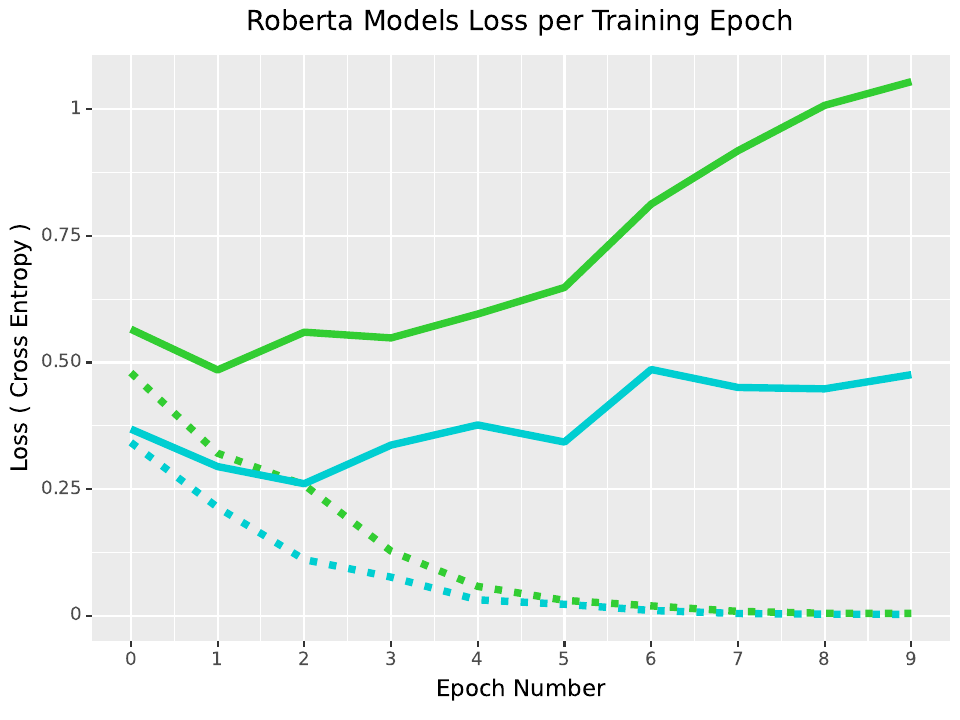} 
         &
     \includegraphics[width=0.65\columnwidth, scale=0.6]{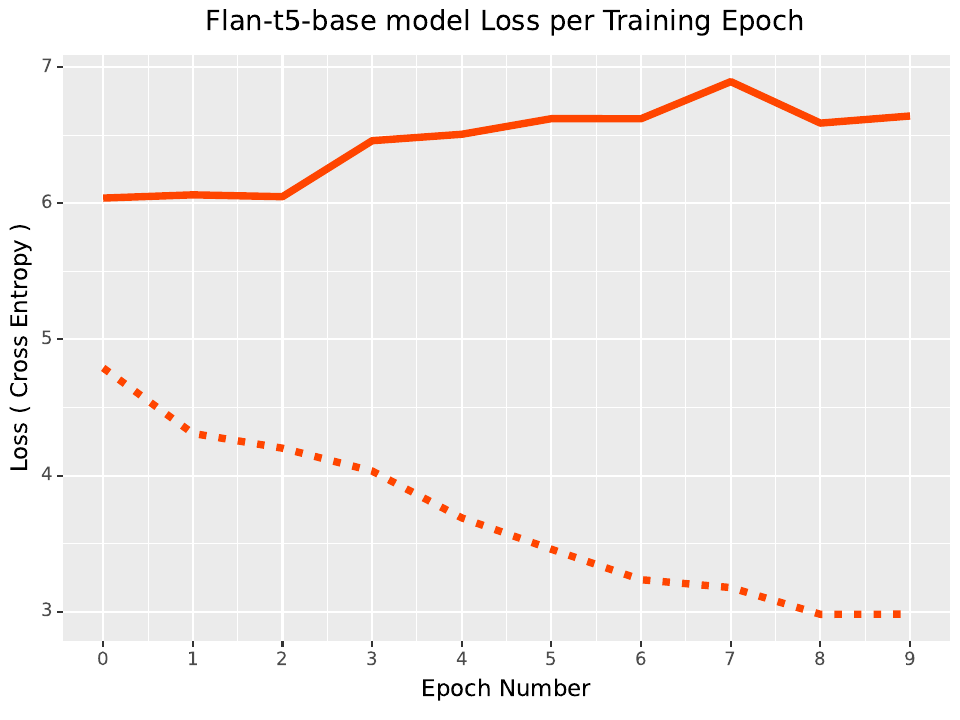} 
         &
     \includegraphics[width=0.65\columnwidth, scale=0.6]{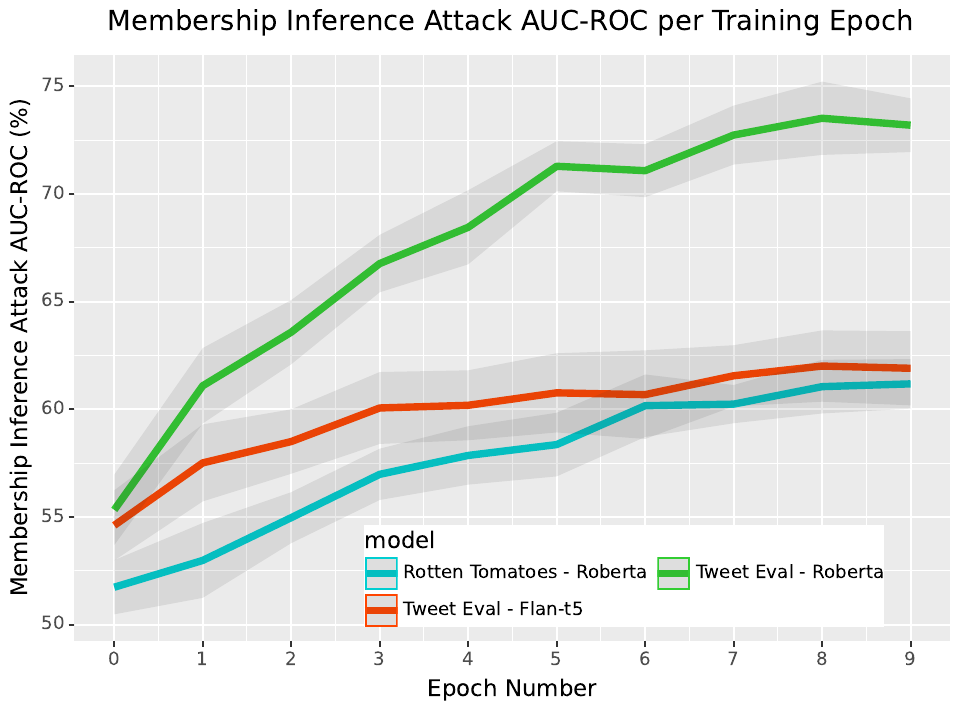} 
    \end{tabular}
\caption{Effect of number of training iterations on MIA success rate. \\Left and middle plots - Dotted lines indicate training loss, full lines - test loss.}
\label{fig:overiftting}
\end{figure*}


\begin{figure}[ht]
    \centering
    \includegraphics[width=0.95\columnwidth, scale=0.7]{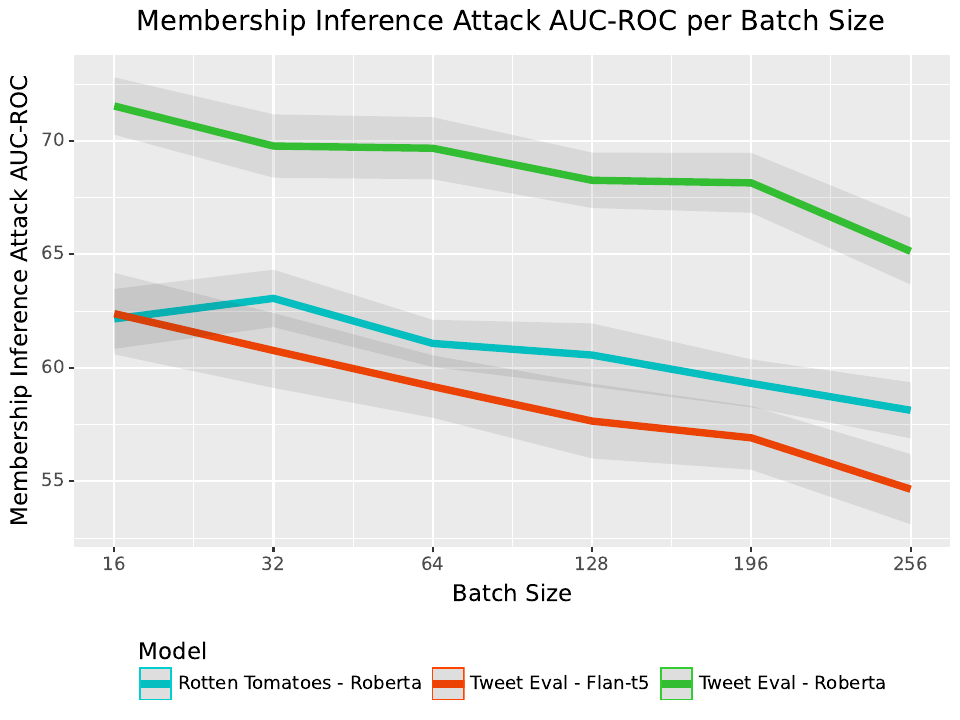} 
    \caption{Effect of batch size on MIA success rate}
    \label{fig:batch_size}
\end{figure}


We start by exploring the effect of overfitting on LLMs by examining the over-training factor. Over-training can be caused either by the number of training epochs, or by the number of gradient updates, influenced by the batch size.

\subsubsection{Number of training epochs}
\label{sec:overfiting}
First, we trained Roberta-base and Flan-t5-base models with a varying number of epochs, using a fixed batch size.
In Figure~\ref{fig:overiftting}, we plot three metrics as a function of the number of training epochs: (1) the MIA AUC-ROC (2) the cross-entropy loss on the training set (3) the cross-entropy loss on the test set.
We observe a clear trend of increased vulnerability to MIA (right plot) with the number of training epochs. In the \textit{Roberta-base Tweet Eval} model, this trend is especially pronounced, exhibiting an increasing polynomial behavior. Nonetheless, for the \textit{Roberta-base Rotten Tomatoes} and \textit{Flan-t5-base Tweet Eval} models, the MIA vulnerability stops increasing after the sixth epoch, which is also reflected in the left and middle plots by the non increasing test loss after this epoch.



\subsubsection{Batch size}
\label{sec:batch_size}
After evaluating the MIA vulnerability caused by repeated exposure of the data samples to the model during training, here we analyze the effect of the number of gradient updates.
In our experiments we trained models with increasing batch sizes, while keeping the number of training epochs constant, which results in fewer gradient updates.
All models were trained over ten epochs, and we varied the learning rate such that all the models reach roughly the same test accuracy, $~89\%$, $~79\%$, and $~73\%$ for \textit{Rotten Tomatoes - Roberta}, \textit{Tweet Eval - Roberta} and \textit{Tweet Eval - Flan t5}, respectively.

In Figure~\ref{fig:batch_size}, we plot the training batch size, ranging from 16 to 256, against the MIA AUC-ROC.
The figure presents a clear trend, indicating that as the batch size increases, the vulnerability to MIA decreases. 
This phenomenon can be traced back to the typical procedure used to train a neural network, which consists of averaging the batch samples' gradients and using the result to update the model weights.
When more samples' gradients are averaged, the knowledge infused into the weights is more general, and contributes less accurate information about specific samples (i.e., less memorization).
Since the success of MIA is partially attributed to the memorization of specific samples, using a larger batch size in the training process can increase the model's robustness to these attacks.

\subsection{Model Size} 
\label{sec:model_size}

\begin{figure*}[ht]
    \centering
    \begin{tabular}{cc}
        \includegraphics[width=0.9\columnwidth]{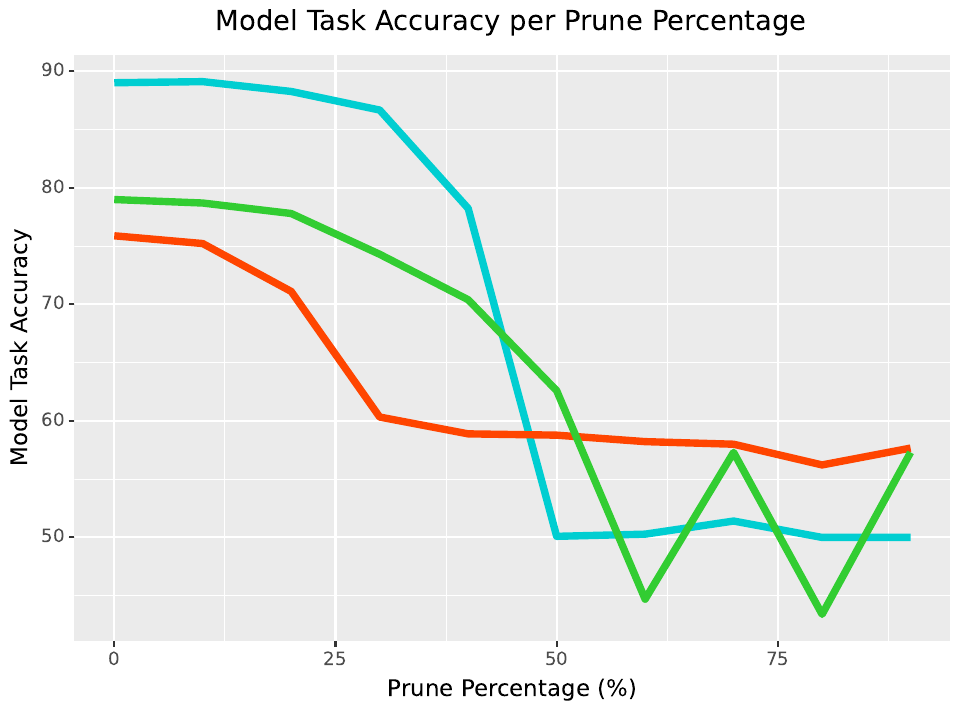}  &
          \includegraphics[width=0.9\columnwidth]{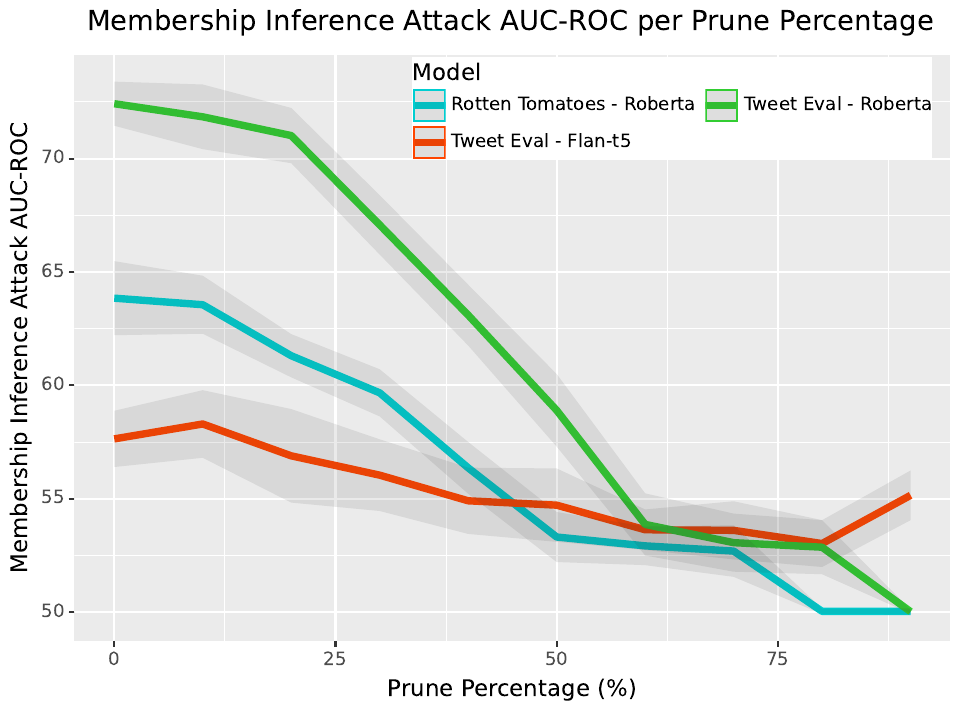}\\       
    \end{tabular}
\caption{Effect of pruning on MIA success rate}
\label{fig:pruning}
\end{figure*}

In this subsection, we explore the effect of the model's size on its vulnerability to MIA, which touches on the over-parametrization factor.
Throughout this section, we discuss two types of mechanisms to reduce model size commonly used for language models: (1) reducing the model size after training and (2) reducing the model size by training a smaller number of parameters. This reduction can also be beneficial for non-privacy related reasons, such as faster inference time, reduced memory consumption and reduced carbon footprint.

\subsubsection{Reducing model size after training}
One approach is to reduce the model's size after training has been completed in an attempt to remove some of the inadvertent memorization that may have taken place during training. 

A straightforward approach to reducing model size is to employ pruning.
Pruning is the process of deleting some of a model's weights according to a given criterion. Unstructured pruning involves deleting model weights with a low absolute value. Structured pruning removes complete modules (such as neurons or heads in transformers) with low $L_1$ norm.

In Figure~\ref{fig:pruning}, we explore the effect of removing increasing amounts of weights from our Roberta-base and Flan-t5-base models fine-tuned on the datasets described in Section~\ref{sec:setup}. 
The weights were removed from the models using unstructured pruning applied to all layers, ranging from the removal of $10\%$ up to $90\%$ of the models' total number of weight parameters. This is equivalent to removing $12M$-$108M$ and $24M$-$222M$ weight parameters from the Roberta-base and Flan-t5-base models, respectively.

As expected, we observe a clear trend indicating that as the percentage of pruned (removed) weights increases, the vulnerability to MIA decreases (right plot), which is especially strongly pronounced for the \textit{Tweet Eval - Roberta} model. However, the left plot in this Figure indicates that unstructured pruning can also have a dire effect on the model's main task accuracy.
Whenever the MIA AUC-ROC drops below $60\%$ for the Roberta models, the task accuracy decreases by more than $20\%$. This shows that pruning is not an effective countermeasure when employed in this manner.

Notably, the fluctuations in the accuracy plots appear to be caused by the fact that after pruning more than $50\%$ of the weights, the models' ability to correctly classify samples is completely degraded and rendered equivalent to outputting a random class. For example, in the \textit{Tweet Eval - Roberta} model, which contains two classes, this manifests as the line jumping around the class ratio point, which is 60\% for the \textit{Tweet Eval} dataset.

Employing structured or unstructured pruning is a trivial approach to reducing the model size, but in the context of MIA, more sophisticated methods have been suggested.
\textit{GAP pruning}~\cite{wang2022analyzing} is a MIA mitigation technique that utilizes pruning along with a special loss function during the fine-tuning process.
In each fine-tuning epoch, the model weights are pruned (using unstructured pruning) to reduce the model size, resulting in the removal of part of the information infused into the model during the training epoch. Since a common feature used by MIA attacks is the entropy of the model’s outputs, \textit{GAP pruning} also employs a loss function that minimizes the variation between the outputs, by enforcing a small difference between the maximal and minimal output values. Formally:
\begin{equation}
    \mathcal{R} = \lambda \cdot ( f(x)_{max} - f(x)_{min})
\end{equation}
where $f$ is the model, $x$ is the input sample, and $\lambda$ is a hyper-parameter controlling the importance of this loss component compared to the main loss function.
Results demonstrating the performance of \textit{GAP pruning}, as well as simple unstructured pruning, can be seen in Table~\ref{tab:size}.

\begin{table*}[ht]
\caption{Effect of reducing the model size on MI attacks success rate. \\Number in parentheses represents standard deviation across 20 experiments.}
\label{tab:size}
\begin{tabular}{cc}
\centering

\resizebox{0.5\textwidth}{!}{  

\begin{tabular}[t]{lc|cccccc}
\hline
\multicolumn{1}{l}{} & \multicolumn{1}{l|}{}                                            & \multicolumn{3}{c}{Rotten Tomatoes - Roberta}                                                                                                                         & \multicolumn{3}{c}{Tweet Eval - Roberta}                                                                                                                              \\ \hline
Model                & \begin{tabular}[c]{@{}c@{}}Trainable\\   Parameters\end{tabular} & \begin{tabular}[c]{@{}c@{}}MIA\\  AUC-ROC\end{tabular} & \begin{tabular}[c]{@{}c@{}}TPR@\\ FPR1\%\end{tabular} & \begin{tabular}[c]{@{}c@{}}Model \\ Acc\end{tabular} & \begin{tabular}[c]{@{}c@{}}MIA\\  AUC-ROC\end{tabular} & \begin{tabular}[c]{@{}c@{}}TPR@\\ FPR1\%\end{tabular} & \begin{tabular}[c]{@{}c@{}}Model \\ Acc\end{tabular} \\ \hline
Base                 & 125M     &      63.00(1.4)   &   1.22(0.6)   &      89.30    &   79.81(0.9)   &        2.58(0.9)     &     79.11  \\
Pruning             & 85M       &  59.64(1.0)    &  1.48(0.8)   &    85.67 &  67.07(1.3) & 1.41(0.8) & 69.30  \\
Distilled            & 82.8M   &       67.27(1.3)          &               1.42(0.7)     &         86.11          &    73.56(1.0)            &    2.32(0.9)     &  78.15       \\
GAP Pruning            & 73M     &      57.80(1.2)            &      1.49(0.8)      &      84.24     &          58.04(1.7)      &      1.42(0.5)   &           74.52        \\
LoRA                 & 1.5M      &      58.02 (1.1)         &       1.38(0.5)    &    88.18        &   58.20(0.9)   &  1.47(0.8)   &   79.83     \\
Distilled+LoRA       & 1.3M  &     59.65(1.5)   &      1.32(0.6)   &      86.58     &  60.16(1.9) &    1.41(0.7)  &      79.26                                                \\ \hline
\end{tabular}    
}

     & 

\resizebox{0.325\textwidth}{!}{  

\begin{tabular}[t]{lc|ccc}
\hline
\multicolumn{1}{l}{} & \multicolumn{1}{l|}{}                                            & \multicolumn{3}{c}{Tweet Eval - Flan-t5}                                                                                                                              \\ \hline
Model                & \begin{tabular}[c]{@{}c@{}}Trainable\\   Parameters\end{tabular} & \begin{tabular}[c]{@{}c@{}}MIA\\  AUC-ROC\end{tabular} & \begin{tabular}[c]{@{}c@{}}TPR@\\ FPR1\%\end{tabular} & \begin{tabular}[c]{@{}c@{}}Model \\ Acc\end{tabular} \\ \hline
Base      &     247.5M      &       57.62(1.5)     &     1.38(0.6)          &   75.88  \\
Pruning    & 172M  &  56.0(1.5)   & 2.51(1.5)   &  60.33 \\
Small      &    76.9M      &        54.33(1.4)     &   1.05(0.4)    &  70.0                                                         \\
LoRA        &   0.88M      &          56.31(1.5)     &   1.63(0.7)    &     74.00            \\
Small+LoRA     &  0.34M    &   53.81(1.2)    &  1.21(0.7)   &     69.66                                                          \\
Prompt Tuning        & 30K   &   55.09(1.4)     &   1.63(0.5)   & 65.66  \\ \hline
\end{tabular}}
     
\end{tabular}

\end{table*}

For \textit{GAP pruning}, we used a prune percentage of $40\%$ and a $\lambda$ hyper-parameter value of $1.5$, which we determined using the model's performance on the validation set. For unstructured pruning, we present results for the removal of 30\% of the weights, since for most of our models this percentage strikes a good balance between the MIA AUC-ROC and the model task accuracy ($(ACC + (1-AUC)) /2$).

\subsubsection{Training less parameters}
Since fine-tuning all of the model's parameters in LLMs can require a significant amount of computational power, a common procedure when fine-tuning a model is to train a lighter version of the model or to train only part of it.
The privacy motivation of using such methods is similar to before, but unlike the previous methods which act as post-hoc mechanisms, these approaches limit the information transferred to the model in the first place.

\textit{Knowledge distillation}~\cite{hinton2015distilling} is a technique for transferring the knowledge from an ensemble of models or a large teacher model into a smaller student model. This technique is often used with language models to create a smaller version of a large pre-trained model such as Roberta or GPT2 to facilitate deployment and inference on limited hardware.
For example, the distilled version of Roberta-base has 6 transformer blocks instead of 12 in the original model, and only exhibits a small gap in task performance. In our evaluation, we check if this difference of $\sim$40M in the number of weight parameters (6 transformer blocks) from Roberta-base renders it less susceptible to MIA.

Since there is no publicly available distilled version of Flan-t5-base, we use a small version of Flan-t5 containing 76.6M parameters, comparing to the base version which contains 247.5M parameters.

\textit{Low Rank Adaptation (LoRA)}~\cite{hu2021lora} - this recent work shows that pre-trained language models require minimal adaptation to perform a new task. Low rank adaptation is a technique that allows to fine-tune a pre-trained language model for a new task by inserting small adaptation layers into the model.
Each model layer is coupled with a pair of two small matrices, whose multiplication results in the change required in a given layer to fit it for a new task.
Formally, given a fully connected layer containing a weight matrix $A\in\mathcal{R}^{N\times N}$, \textit{LoRA} couples it with two smaller matrices $B_r\in\mathcal{R}^{N\times R}$ and $B_l\in\mathcal{R}^{R\times N}$. Given an input $x\in\mathcal{R}^{N\times 1}$, the adaptation is performed by:
\begin{equation}
    \bar{A}(x) = (A + B_rB_l)x
\end{equation}
where $\bar{A}$ is the layer operation after inserting the adaptation matrices.

During the fine-tuning process, the original weight matrices are frozen (not trained) and only the adapter layer matrices are modified.
The advantage of using this method stems from the fact that for successful fine-tuning, the inner dimension of the weight matrices $B_r$ and $B_l$ can be very small, for example $8$, resulting in the training of only a fraction of the original model weight parameters (1.5M for Roberta-base instead of 120M).
The implications of \textit{LoRA} on the privacy of language models are significant, since it diminishes the effective capacity of the model to memorize the training set.
Our evaluation (as presented in Table~\ref{tab:size}) demonstrates the effect of using \textit{LoRA}, as well as the combination of \textit{LoRA} with a distilled/small version of the model, using an inner dimension $R$ of $8$.

For generative LLMs, aside from \textit{LoRA}, there is another parameter-efficient tuning method called \textit{Prompt-tuning}, which in some cases can adapt an LLM for a new task while training only thousands of parameters (e.g., $30K$).
Prompt-tuning works by extending the LLM vocabulary with a few artificial tokens, which are appended to each text inputted to the model.
When using this method for tuning, similarly to \textit{LoRA}, the original model weights are frozen, while only the new tokens' embedding weights are trained.
The new tokens' embeddings affect the operation of the frozen LLM and help to adapt its outputs to the new task.

Table~\ref{tab:size} summarizes the MIA AUC-ROC, MIA TPR@FPR1\%, and main task accuracy of the models presented in this section (\ref{sec:model_size}).
Although model size is considered one of the most influential factors on training data memorization and MIA vulnerability, we observe that there are some other important considerations that should be addressed when constructing a MIA-robust model.

Simply reducing the model size after training using unstructured pruning does not gain significant protection from MIA, and can seriously degrade the model's main task accuracy. Nevertheless, when pruning is employed as part of the training process (e.g., \textit{GAP pruning}), the model task accuracy slightly improves while still managing to provide protection against MIA.

Employing knowledge distillation, even using a much smaller replica of the model ($\frac{2}{3}$ of the original model size), did not provide any form of mitigation against MIA for the Roberta models, and even increased its vulnerability to MIA in the \textit{Rotten Tomatoes - Roberta} case.
Interestingly, in the \textit{Tweet Eval - Flan-t5} case, where we used a smaller model that is not the product of a distillation process, it did provide meaningful mitigation (3.3 difference in AUC-ROC, and 0.3 in TPR@FPR1\%). This may be due to the fact that this model is substantially smaller than its corresponding base model (almost $\frac{1}{3}$ of its size). Moreover, it is important to note that in this case this also resulted in a significant decrease in task accuracy ($5.8\%$ less than the base Flan-t5 model).

With \textit{LoRA}, we were able to reduce the number of model trainable parameters to a small fraction of those in the original models ($1.5M$ for Roberta-base and $0.88M$ for Flan-t5-base).
This very small amount of trainable parameters provides fine-tuned models with a strong defense against MIA.
Throughout our evaluations, all models fine-tuned using \textit{LoRA} resulted in high reduction in MIA vulnerability, reaching a maximal MIA AUC-ROC of $58.2\%$.
These results appears to be amplified when combining \textit{LoRA} with a smaller \textit{Flan-t5} model, which further reduces both MIA AUC-ROC and MIA TPR@FPR1\%.
However, the application of \textit{LoRA} to the \textit{Distilled} Roberta models did not lead to an improvement over the application of \textit{LoRA} to the Roberta-base models in terms of MIA vulnerability, and only provided an improvement over the \textit{Distilled} models.

With \textit{Prompt-tuning}, the \textit{Tweet Eval - Flan-t5} model did reduce it's vulnerability to MIA, even compared to the \textit{LoRA} model, however this approach may damage the task accuracy, as is the case with the \textit{Tweet Eval} dataset (see Table~\ref{tab:flan_table} in the appendix for \textit{Prompt-tuning} results on Rotten Tomatoes).

Overall, this evaluation shows that model size is an important factor when protecting language models against MIA.
However, from Figure~\ref{fig:pruning} and from the results in Table~\ref{tab:size}, we deduce that to completely mitigate MIA, the model size should be reduced significantly (by more than $50\%$).

\subsection{Differential Privacy} 
\label{sec:DP}
\begin{figure*}[ht]
    \centering
    \begin{tabular}{cc}
             \includegraphics[width=0.9\columnwidth]{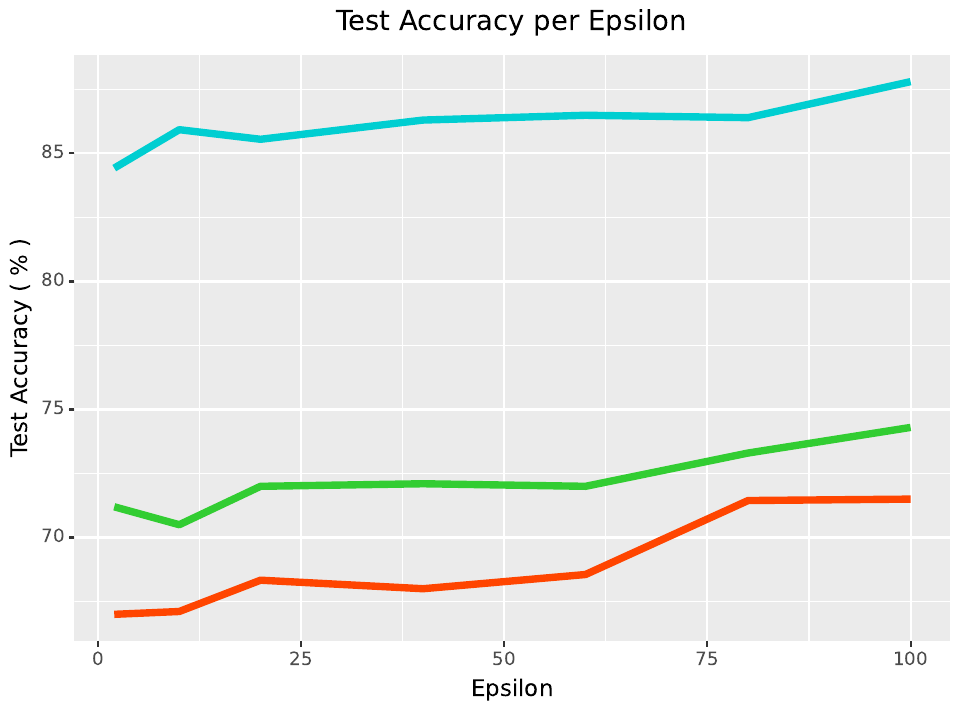} &
             \includegraphics[width=0.9\columnwidth]{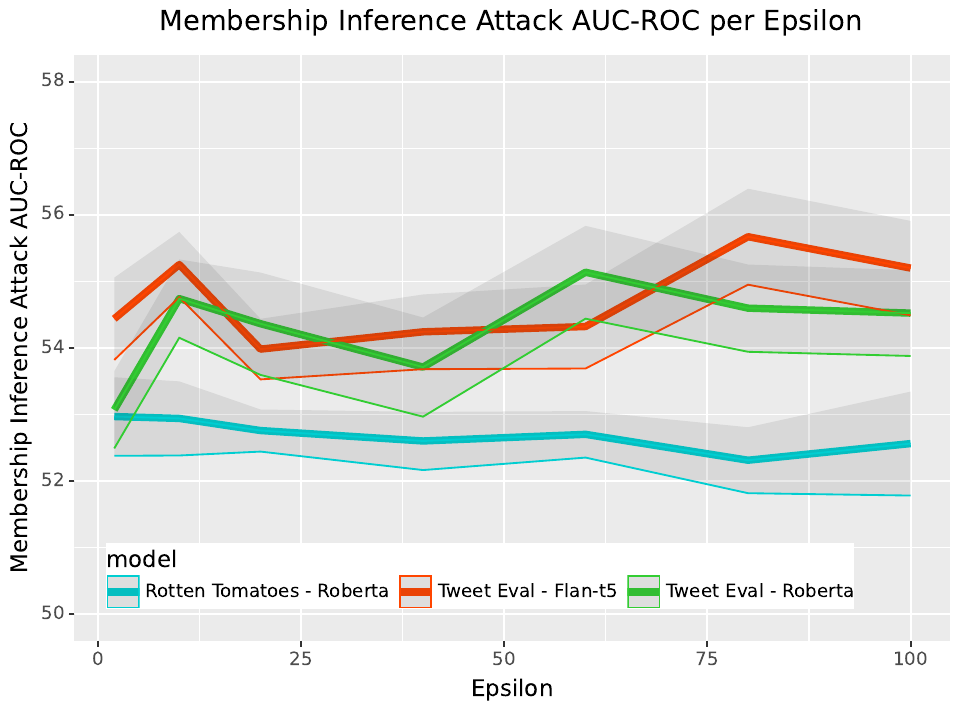} 
         \\       
    \end{tabular}
\caption{Effect of $\epsilon$ in DP-SGD on MIA success rate for the Rotten Tomatoes dataset}
\label{fig:epsilon}
\end{figure*}
\begin{table*}[t]
\caption{MIA AUC-ROC of different DP hyper-parameter combinations}
\label{tab:DP_params}
\centering

\begin{tabular}{ccc}

\resizebox{0.3\textwidth}{!}{
\begin{tabular}{cccc}
\hline
\multicolumn{4}{c}{Rotten Tomatoes - Roberta}                                                                                           \\ \hline
\multicolumn{1}{c|}{}       & \multicolumn{3}{c}{Max Grad Norm (C)}                                                                     \\ \hline
\multicolumn{1}{c|}{Epochs} & 0.1                               & 0.09                              & 0.08                              \\ \hline
\multicolumn{1}{c|}{3}      & \cellcolor[HTML]{FCFCFF}52.77514  & \cellcolor[HTML]{F8696B}53.357447 & \cellcolor[HTML]{FBC6C9}52.990468 \\
\multicolumn{1}{c|}{6}      & \cellcolor[HTML]{5A8AC6}51.944069 & \cellcolor[HTML]{769DCF}52.08879  & \cellcolor[HTML]{FBB5B8}53.058129 \\
\multicolumn{1}{c|}{10}     & \cellcolor[HTML]{D9E3F2}52.598859 & \cellcolor[HTML]{AAC2E2}52.35853  & \cellcolor[HTML]{FCFCFF}52.776105 \\ \hline
\end{tabular}
}
    
         & 
 \resizebox{0.3\textwidth}{!}{        
\begin{tabular}{clll}
\hline
\multicolumn{4}{c}{Tweet Eval - Roberta}                                                                                                \\ \hline
\multicolumn{1}{c|}{}       & \multicolumn{3}{c}{Max Grad Norm (C)}                                                                     \\ \hline
\multicolumn{1}{c|}{Epochs} & \multicolumn{1}{c}{0.1}           & \multicolumn{1}{c}{0.09}          & \multicolumn{1}{c}{0.08}          \\ \hline
\multicolumn{1}{c|}{3}      & \cellcolor[HTML]{5A8AC6}53.166148 & \cellcolor[HTML]{9DB9DD}53.543312 & \cellcolor[HTML]{BDD0E9}53.720018 \\
\multicolumn{1}{c|}{6}      & \cellcolor[HTML]{FCFCFF}54.065019 & \cellcolor[HTML]{FCE8EB}54.301987 & \cellcolor[HTML]{CCDAEE}53.801506 \\
\multicolumn{1}{c|}{10}     & \cellcolor[HTML]{FAA1A4}55.105737 & \cellcolor[HTML]{FCE4E7}54.340882 & \cellcolor[HTML]{F8696B}55.741754 \\ \hline
\end{tabular}
}                 
         &

\resizebox{0.3\textwidth}{!}{
\begin{tabular}{clll}
\hline
\multicolumn{4}{c}{Tweet Eval - Flan-t5}                                                                                                \\ \hline
\multicolumn{1}{c|}{}       & \multicolumn{3}{c}{Max Grad Norm (C)}                                                                     \\ \hline
\multicolumn{1}{c|}{Epochs} & \multicolumn{1}{c}{0.1}           & \multicolumn{1}{c}{0.09}          & \multicolumn{1}{c}{0.08}          \\ \hline
\multicolumn{1}{c|}{3}      & \cellcolor[HTML]{AAC2E2}54.649886 & \cellcolor[HTML]{FBCACD}55.291712 & \cellcolor[HTML]{5A8AC6}54.440896 \\
\multicolumn{1}{c|}{6}      & \cellcolor[HTML]{FCFCFF}54.863819 & \cellcolor[HTML]{F8696B}56.109212 & \cellcolor[HTML]{FCEEF1}54.986468 \\
\multicolumn{1}{c|}{10}     & \cellcolor[HTML]{D8E2F2}54.7701   & \cellcolor[HTML]{FAA4A7}55.610222 & \cellcolor[HTML]{CDDBEE}54.742191 \\ \hline
\end{tabular}    
}       
         \\
         
    \end{tabular}
\end{table*}
\begin{table*}[t]
\caption{Comparison between DP-SGD (full fine-tuning) and DP-LoRA. \\Number in parentheses represents standard deviation across 20 experiments.}
\label{tab:dp_vs_dp_lora}
\centering
\begin{tabular}{l|cccccc|ccc}
\hline
                      & \multicolumn{6}{c|}{Roberta - base}                                                                                                                                                                                                                                                                                                       & \multicolumn{3}{c}{Flan-t5-base}                                                                                                                                    \\ \hline
                      & \multicolumn{3}{c}{Rotten Tomatoes}                                                                                                                                 & \multicolumn{3}{c|}{Tweet Eval}                                                                                                                                     & \multicolumn{3}{c}{Tweet Eval}                                                                                                                                      \\ \hline
                      & \begin{tabular}[c]{@{}c@{}}MIA\\ AUC-ROC\end{tabular} & \begin{tabular}[c]{@{}c@{}}TPR@\\ FPR1\%\end{tabular} & \begin{tabular}[c]{@{}c@{}}Model\\ Acc\end{tabular} & \begin{tabular}[c]{@{}c@{}}MIA\\ AUC-ROC\end{tabular} & \begin{tabular}[c]{@{}c@{}}TPR@\\ FPR1\%\end{tabular} & \begin{tabular}[c]{@{}c@{}}Model\\ Acc\end{tabular} & \begin{tabular}[c]{@{}c@{}}MIA\\ AUC-ROC\end{tabular} & \begin{tabular}[c]{@{}c@{}}TPR@\\ FPR1\%\end{tabular} & \begin{tabular}[c]{@{}c@{}}Model\\ Acc\end{tabular} \\ \hline
DP-SGD($\epsilon$=2)  & 52.59(1.0) & 1.01(0.4)   & 87.05  & 54.34(1.3) & 1.26(0.6)  & 72.4 & 54.98(1.4) & 1.20(0.5) & 73.11  \\
DP-LoRA($\epsilon$=2) &    52.61(1.0) &  1.20(0.5)  & 87.42 & 53.58(1.3)    &   0.88(0.4)         &    74.0 &    55.10(1.0)  & 1.51(0.9)  &  71.11 \\ \hline
\end{tabular}
\end{table*}

\textit{DP-SGD} \cite{abadi2016deep} is an application of DP for neural network training that guarantees the privacy of the output of the training procedure, i.e., the model weights.
With the weights guaranteed to be private, \textit{DP-SGD} and it's variations should provide protection against privacy attacks in both the black-box scenario, when the adversary analyzes the model's outputs, and the white-box scenario, when the model weights are available to the attacker. DP-SGD utilizes two concepts previously discussed in this paper: (1) varying the model's exposure to training samples by clipping and noising each sample gradient, and (2) limiting the number of training iterations (as a larger number of gradient updates results in a higher privacy budget $\epsilon$). In this paper, we refer to DP-SGD as the application of this approach to the model's fine-tuning process, i.e., when performing full fine-tuning of the model.

This section provides three types of evaluations of the robustness to MIA gained by using DP algorithms: (1) exploring the effect of $\epsilon$ on MIA robustness ; (2) exploring the effect of different DP hyper-parameters on MIA robustness; and (3) a comparison between \textit{DP-SGD} full fine-tuning and another variant that applies DP to the \textit{LoRA} fine-tuning method, called \textit{DP-LoRA}.

\subsubsection{Effect of $\epsilon$ on the vulnerability to MIA}
$\epsilon$, also known as the privacy budget, is a quantitative measure of the model's privacy when employing DP. When $\epsilon$ is smaller, it is expected that the model's privacy leakage will be smaller.
However, the definition of $\epsilon$ is theoretical, and not based on actual vulnerability to privacy attacks.
This evaluation sets out to demonstrate the effect of $\epsilon$ on the model's effective robustness to MIA.

We evaluate the performance of Roberta-base and Flan-t5-base models fine-tuned on the \textit{Rotten Tomatoes} dataset and \textit{Tweet Eval} datasets using DP-SGD with varying values of $\epsilon$. We used a batch size of 250 and matching differential privacy and training parameters.

Our results, presented in Figure~\ref{fig:epsilon}, show an expected trend where increasing the privacy budget (larger $\epsilon$) improves the model's accuracy. 
However, there is only a marginal increase in MIA vulnerability (right plot), even as $\epsilon$ reaches a large, theoretically meaningless value. In fact, in accordance with the differential privacy definition (see Section~\ref{sec:background}), $\epsilon$ values larger than 1 do not provide a meaningful probability bound.
Nevertheless, our results suggest that even using a large $\epsilon$ can be a viable alternative for providing adequate empirical privacy while still maintaining acceptable model accuracy, in situations where training a model with a small $\epsilon$ is challenging. 

\subsubsection{Exploration of the hyper-parameter space of DP-SGD}
The \textit{DP-SGD} algorithm has several hyper-parameters that dictate the resulting privacy budget $\epsilon$ . As discussed before in Section~\ref{sec:DP}, $\epsilon$ is based on a theoretical probability bound, and not on robustness to privacy attacks. However, similar to other hyper-parameters used for model training, the hyper-parameters for \textit{DP-SGD} may have a large effect on the model's performance, as well as on its effective privacy.
As different combinations of DP hyper-parameters can lead to the same $\epsilon$ value, the privacy implications must be considered when choosing them.

In this evaluation we set out to explore the effect of different combinations of hyper-parameters on MIA robustness. Although there are several hyper-parameters for DP, here we have focused on the max gradient norm $C$ and the number of training epochs.

For this evaluation, we fined-tuned our models on the \textit{Rotten Tomatoes} and \textit{Tweet Eval} datasets using several combinations of hyper-parameters corresponding to an $\epsilon$ value of $2$. We used a fixed batch size of $250$, and inferred the required additive gradient noise $\sigma$ from the other hyper-parameters.
The learning rate for training was set empirically so as to maximise the model's performance on the validation set, and all models resulted in roughly the same accuracy.

In Table~\ref{tab:DP_params}, we present the average MIA AUC-ROC for each of the models, and color the cells based on attack vulnerability (red being more vulnerable and blue less vulnerable).
The numbers in the tables exhibit relatively high variability between the different combinations.
Notably, for each model we observe a difference of almost $2$ AUC percentage points between different hyper-parameter combinations, even though these different combinations lead to the same theoretical privacy guarantee ($\epsilon$=2).
We identify a common pattern between the different model types, where models trained using a maximal gradient norm of $0.1$ tend to have better AUC-ROC scores.
This highlights the importance of the additive noise ($\sigma$), which needs to be amplified to compensate for the relatively high maximal gradient norm.

In conclusion, not all DP hyper-parameters lead to equal robustness to MIA, even when all examined combinations correspond to the same $\epsilon$ value.
In practice, when developing a model, the choice of these DP hyper-parameters should be made while considering both robustness to privacy attacks and model performance.

\subsubsection{DP-SGD vs DP-LoRA}
A notable variation of \textit{DP-SGD} designed for fine-tuning LLMs is \textit{DP-LoRA}~\cite{yu2021differentially}, which applies modified gradient updates only to the adaptation layers (\textit{LoRA} weights), making the application of DP to LLMs easier with limited decrease in task accuracy.

In Table~\ref{tab:dp_vs_dp_lora}, we compare the performance and MIA susceptibility of the \textit{DP-LoRA} and \textit{DP-SGD} (full fine-tuning) algorithms. Specifically, we evaluate these algorithms at a privacy budget of $2$, utilizing a batch size of $250$ and hyper-parameters that lead to the best model performance on the validation sets.

Our findings demonstrate that while \textit{DP-LoRA} does not  enhance MIA resilience compared to \textit{DP-SGD}, it facilitates a less complex and faster training procedure, leading to increased model accuracy (of up to 2\%).
However, it is important to note that employing \textit{DP-LoRA} relies on the assumption that the pre-training of the model, i.e., training of all the backbone weights, was performed on non-private data.
In conclusion, based on our assessment, \textit{DP-LoRA} emerges as the preferred choice for mitigating MIA risk, provided that such prerequisites are met.

DP can technically also be applied to other methods of parameter-efficient fine-tuning, such as \textit{Prompt-tuning}. Although during this effort we were not able to tune a usable Flan-t5-base model for the \textit{Tweet Eval} dataset using DP-prompt-tuning, we did, however, manage to do so for the \textit{Rotten Tomatoes} dataset, as shown in the Appendix (Table ~\ref{tab:flan_table}).

\subsection{Results Summary}
\label{sec:results_summary}
\begin{figure*}[ht!]
    \begin{tabular}{ccc}
         \includegraphics[width=0.6\columnwidth]{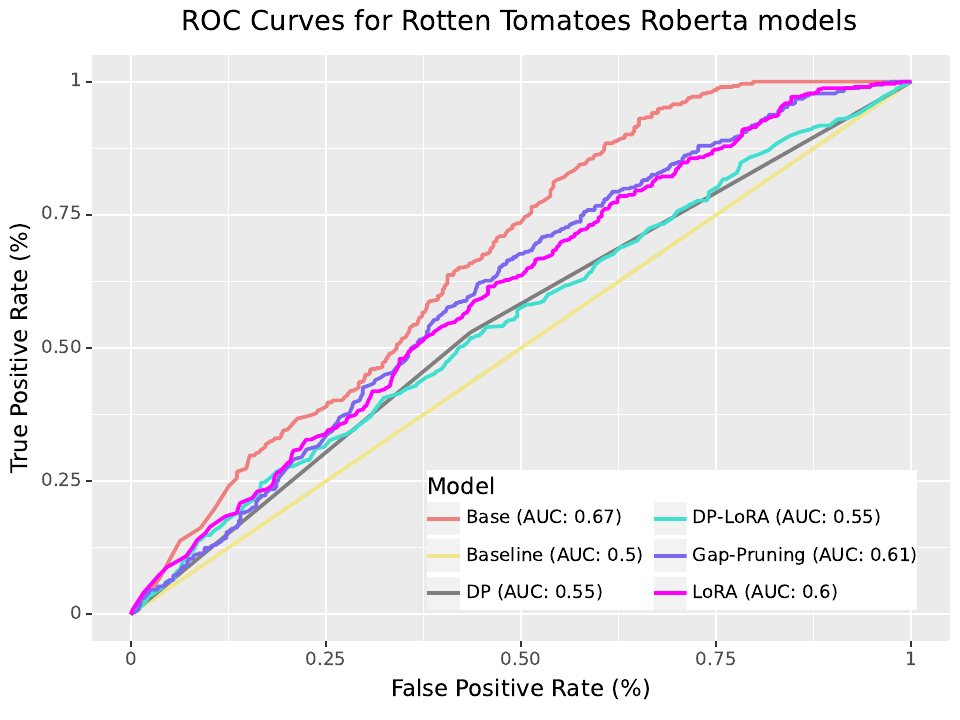} &
         \includegraphics[width=0.6\columnwidth]{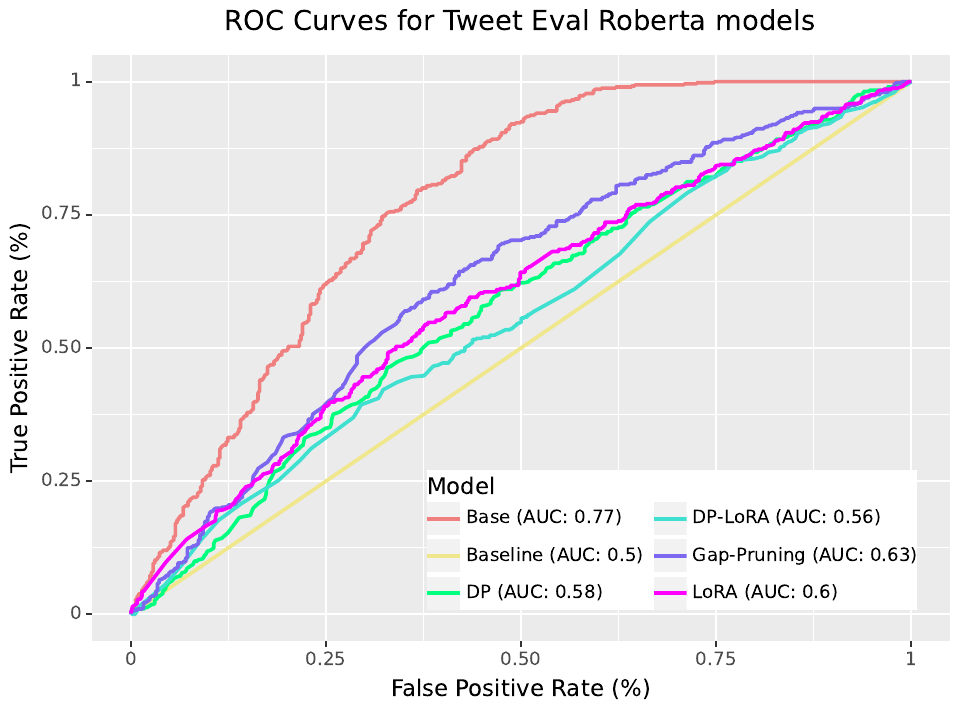}&
        \includegraphics[width=0.6\columnwidth]{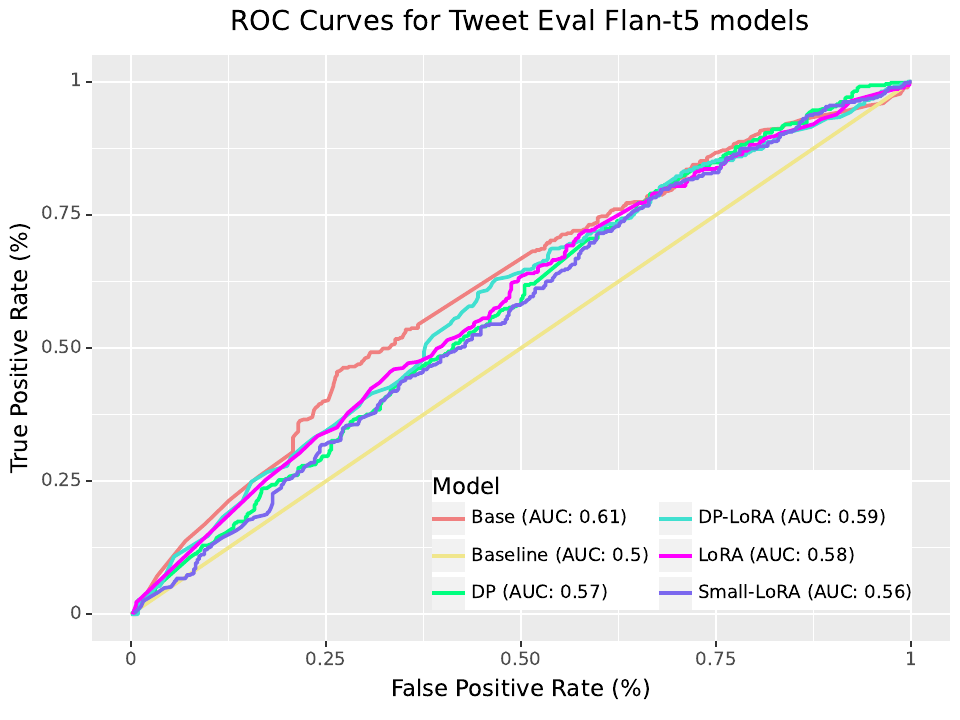}     
    \end{tabular}
    \caption{Complete ROC curves for leading mitigation methods}
    \label{fig:best_methods}
\end{figure*}
In our evaluation, we examined three factors affecting the vulnerability of LLMs to MIA: the number of training iterations, the model size, and the model's exposure to the training samples.
Our analysis reveals several key findings that can be translated into practical guidelines for developing privacy-preserving LLMs.

First, we found that training the model using fewer iterations (larger batch sizes and/or fewer epochs) results in a less vulnerable model. This correlates with a known phenomenon from applying DP to model training, that increasing the number of training iterations causes an increase in the privacy budget ($\epsilon$). This finding directly translates to a general guideline for training more private LLMs.

Second, we showed that fine-tuning LLMs with as few as possible model weight parameters results in increased robustness to MIA. Our use of the \textit{LoRA} technique allows for fine-tuning with as little as $\frac{1}{120}$ of the original number of trainable parameters, significantly improving privacy while minimally degrading the task accuracy.

Third, we found that the combination of \textit{LoRA} with \textit{DP-SGD} results in similar protection from MIA as full fine-tuning using standard \textit{DP-SGD}.
However, the reduced number of weights required to be trained under DP constraints (gradient noising, gradient clipping, and a small number of training iterations) in \textit{DP-LoRA} facilitates a smaller degradation in the model's task accuracy compared to standard \textit{DP-SGD}, which is a notable advantage.

To conclude our evaluation, we gathered the most promising methods from each section and performed a worst-case privacy evaluation, selecting the attack that exhibited the highest AUC-ROC score out of all 20 experiments, and presenting in Figure~\ref{fig:best_methods} the full ROC curves of the different models.

Analyzing these ROC curves enables examining two main characteristics (1) the overall area underneath the curve (AUC) which corresponds to the overall MIA leakage, and (2) the TPR at low FPR values, which represents the ability of the attack to identify member samples with high confidence.
Overall, the \textit{DP-SGD} and \textit{DP-LoRA} curves are closest to the baseline curve and have the lowest AUC-ROC for the Roberta models, and \textit{Small-LoRA} has the best result for the Flan-t5 model. However, it is worth noticing that the \textit{Small-LoRA} model causes a $6\%$ accuracy drop compared to the base model, making it less usable than the $DP$ based methods which only cause a drop of up to $3\%$ in model accuracy.

When examining the area of low FPRs, other methods seem to provide a comparable defense. For example, the \textit{GAP Prune} method for the \textit{Rotten Tomatoes - Roberta} model results in a very flat curve for FPR values of less than $20\%$.

Overall, the DP-based approaches we evaluated seem to offer the best privacy/accuracy tradeoff. 
However, the use of such methods can come at the cost of complex training procedures, and in some cases they may still compromise model performance too much. When such methods are not feasible, our experiments show that \textit{LoRA} fine-tuning provides a reasonable defense against MIA while maintaining comparable model performance to regular fine-tuning. When focusing on low FPRs, \textit{GAP-Prune} can also be a viable approach.

\section{Conclusion}
\label{sec:conclusion}

Large language models have gained significant attention in recent years due to their impressive performance on various natural language processing tasks such as language translation and question answering. However, these models also raise serious concerns about data privacy. The vulnerability of LLMs to privacy attacks, and specifically to membership inference attacks, is a pressing concern for individuals and organizations that rely on these models for various applications.

Although there has been a significant amount of research focused on protecting deep neural networks from membership inference attacks, most of this work has been focused on image classification tasks and did not consider defense strategies specifically adapted for LLMs. In this study, we bridged this gap by performing a comprehensive empirical analysis of the factors that influence the vulnerability of fine-tuned LLMs to membership inference attacks and exploring possible mitigation techniques.

Our results indicate that the effective model size (or the number of trainable parameters) is an important factor in determining the susceptibility of LLMs to membership inference attacks. 
However, other factors such as the training batch size and the number of training iterations are also important when designing a model to be resistant to MIA. We found that batch size has a significant effect on MIA vulnerability, with larger batch sizes providing good protection against this type of attack.

Furthermore, we demonstrated that differential privacy can be successfully applied for fine-tuning LLMs through the use of \textit{DP-LoRA}, which offers a scalable and efficient means of defending against membership inference attacks. In most cases, \textit{DP-LoRA}
was the most effective defense strategy. However, since the use of DP can be complex and may cause a degradation in model performance, the use of \textit{LoRA} fine-tuning can also provide a good balance between model performance and protection from MIA.

Our findings contribute to the growing body of research focused on developing defenses for LLMs against privacy attacks. We believe that our work will motivate further research in this area and help to ensure the safe and responsible use of these powerful models.

\section{Acknowledgments}
This work was performed as part of the NEMECYS project, which is co-funded by the European Union under grant agreement ID 101094323, by UK Research and Innovation (UKRI) under the UK government’s Horizon Europe funding guarantee grant numbers 10065802, 10050933 and 10061304, and by the Swiss State Secretariat for Education, Research and Innovation (SERI).

\bibliographystyle{ACM-Reference-Format}
\bibliography{bib}

\appendix
\section{Appendix}
\subsection{Dataset Size and MIA Vulnerability}
\label{app:dataset_size}
\begin{figure}
    \centering
    \includegraphics[width=0.9\columnwidth]{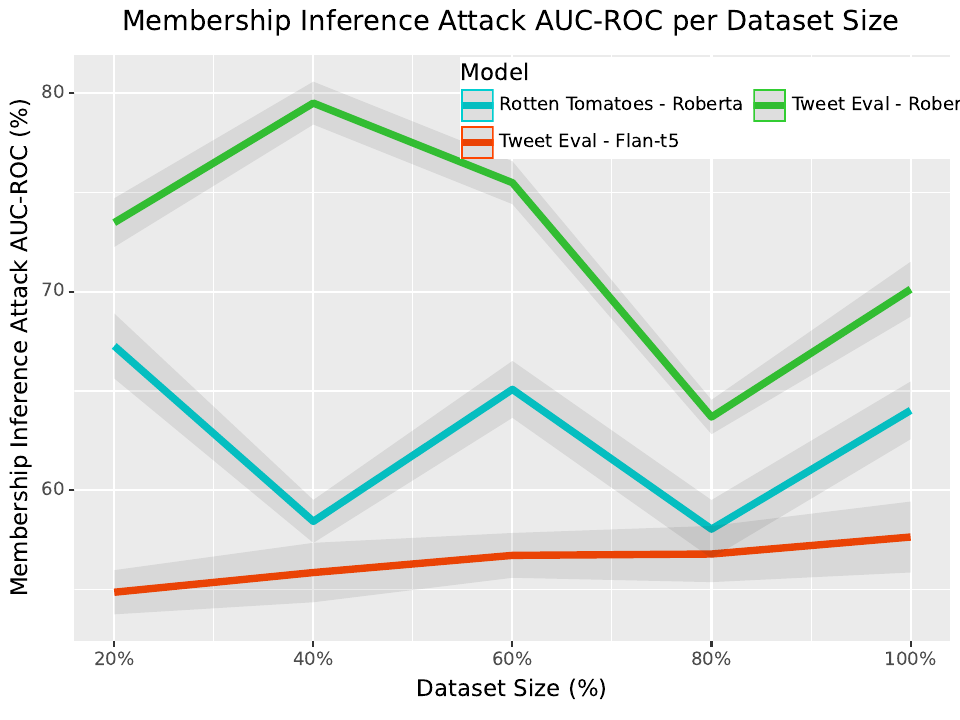}
    \caption{Effect of dataset size on MIA success rate}
    \label{fig:dataset_size}
\end{figure}

In Figure \ref{fig:dataset_size} we analyze the effect of the size of the fine-tuning dataset on MIA vulnerability using the \textit{Rotten Tomatoes} dataset and the \textit{Tweet Eval} dataset. Although the MIA AUC-ROC does present an overall decreasing trend when increasing the dataset size in the Roberta-base models, we do not see the same trend for the Flan-t5 model.

\setcounter{figure}{0}

\subsection{Results for Flan-t5 Models on Rotten Tomatoes}
For the Flan-t5 model fine-tuned on the Rotten Tomatoes dataset (Table \ref{tab:flan_table}), the base model already exhibits quite low vulnerability to MIA, as represented by both AUC-ROC and TPR@FPR1\% being quite close to their respective baseline values. Although there is a slight decrease in these values between the base model and the other alternatives, the differences are quite small and no clear trend is visible as with the other models discussed in this paper.
\begin{table}[ht]
\caption{Results for Rotten Tomatoes - Flan-t5 models}
\label{tab:flan_table}
\begin{tabular}{lc|ccc}
\hline
\multicolumn{1}{l}{} & \multicolumn{1}{l|}{}                                            & \multicolumn{3}{c}{Rotten Tomatoes - Flan-t5}                                                                                                                              \\ \hline
Model                & \begin{tabular}[c]{@{}c@{}}Trainable\\   Parameters\end{tabular} & \begin{tabular}[c]{@{}c@{}}MIA\\  AUC-ROC\end{tabular} & \begin{tabular}[c]{@{}c@{}}TPR@\\ FPR1\%\end{tabular} & \begin{tabular}[c]{@{}c@{}}Model \\ Acc\end{tabular} \\ \hline
Base      &    247M   &   52.93(1.2)   & 1.43(0.8)  &  90.7   \\
DP-SGD        & 247K   &  52.74(0.6)   &   1.14(0.4)    &  87.0 \\

Small              &    172M  &  52.18(0.5) &  1.01(0.7)  &  84.3 \\
LoRA                &  0.85M  &  52.41(0.8) &  1.24(0.6)  &  90.2  \\
DP-LORA       & 0.85M     &  52.72(0.8)   &  1.28(0.6)     &   87.9  \\
Small+LoRA           &   0.34M   & 52.74(0.9) &  1.15(0.5)   &  84.1   \\
Prompt Tuning        & 30K     &  52.18(0.8) &  1.17(0.3)  &  87.6\\
DP Prompt Tuning        & 30K     &  52.74(0.6) &  1.18(0.5)  &  87.0\\
\hline
\end{tabular}
\end{table}

\subsection{Results Summary Table}
Table \ref{tab:summary_table} summarizes the average results (MIA AUC-ROC, TPR@FPR1\%, Model Acc) over 20 runs for the best methods presented in Figure~\ref{fig:best_methods}.
Same as in the worst-case evaluation, here as well the DP based methods emerge as providing the best protection against MIA.
\begin{table*}[hb]
\caption{Results summary for leading mitigation methods}
\label{tab:summary_table}
\centering
\begin{tabular}{l|cccccc|ccc}
\hline
                      & \multicolumn{6}{c|}{Roberta - base}                                                                                                                                                                                                                                                                                                       & \multicolumn{3}{c}{Flan-t5-base}                                                                                                                                    \\ \hline
                      & \multicolumn{3}{c}{Rotten Tomatoes}                                                                                                                                 & \multicolumn{3}{c|}{Tweet Eval}                                                                                                                                     & \multicolumn{3}{c}{Tweet Eval}                                                                                                                                      \\ \hline
                      & \begin{tabular}[c]{@{}c@{}}MIA\\ AUC-ROC\end{tabular} & \begin{tabular}[c]{@{}c@{}}TPR@\\ FPR1\%\end{tabular} & \begin{tabular}[c]{@{}c@{}}Model\\ Acc\end{tabular} & \begin{tabular}[c]{@{}c@{}}MIA\\ AUC-ROC\end{tabular} & \begin{tabular}[c]{@{}c@{}}TPR@\\ FPR1\%\end{tabular} & \begin{tabular}[c]{@{}c@{}}Model\\ Acc\end{tabular} & \begin{tabular}[c]{@{}c@{}}MIA\\ AUC-ROC\end{tabular} & \begin{tabular}[c]{@{}c@{}}TPR@\\ FPR1\%\end{tabular} & \begin{tabular}[c]{@{}c@{}}Model\\ Acc\end{tabular} \\ \hline
Base        &      63.00(1.4)   &   1.22(0.6)   &      89.30    &   79.81(0.9)   &        2.58(0.9)     &     79.11 &  52.93(1.2)   & 1.43(0.8)  &  90.7   \\
LoRA               &      58.02 (1.1)         &       1.38(0.5)    &    88.18        &   58.20(0.9)   &  1.47(0.8)   &   79.83   &  52.41(0.8) &  1.24(0.6)  &  90.2  \\
small + LoRA & n.a. & n.a. & n.a. & n.a. & n.a. &  n.a. & 52.74(0.9) &  1.15(0.5)   &  84.1   \\
GAP Pruning        &      57.80(1.2)            &      1.49(0.8)      &      84.24     &          58.04(1.7)      &      1.42(0.5)   &           74.52      & n.a & n.a. & n.a.  \\
DP-SGD($\epsilon$=2)  & 52.59(1.0) & 1.01(0.4)   & 87.05  & 54.34(1.3) & 1.26(0.6)  & 72.4 & 54.98(1.4) & 1.20(0.5) & 73.11  \\
DP-LoRA($\epsilon$=2) &    52.61(1.0) &  1.20(0.5)  & 87.42 & 53.58(1.3)    &   0.88(0.4)         &    74.0 &    55.10(1.0)  & 1.51(0.9)  &  71.11 \\

\hline
\end{tabular}
\end{table*}

\end{document}